\documentclass[sigconf]{acmart}
\AtBeginDocument{%
  }

\copyrightyear{2026}
\acmYear{2026}
\setcopyright{cc}
\setcctype{by}
\acmConference[MM '26]{Proceedings of the 34th ACM International Conference on Multimedia}{November 10--14, 2026}{Rio de Janeiro, Brazil}
\acmBooktitle{Proceedings of the 34th ACM International Conference on Multimedia (MM '26), November 10--14, 2026, Rio de Janeiro, Brazil}
\acmDOI{10.1145/3767308.3835940}
\acmISBN{979-8-4007-2213-4/2026/11}
\usepackage{multirow}
\usepackage{pifont}

\providecommand{\cmark}{\textcolor{green!60!black}{\ding{51}}}
\providecommand{\xmark}{\textcolor{red!75!black}{\ding{55}}}

\begin{document}

\title{SceneReVis: A Self-Reflective Vision-Grounded Framework for 3D Indoor Scene Synthesis via Multi-Turn RL}

\author{Yang Zhao}
\email{runder1103@sjtu.edu.cn}
\orcid{0009-0009-9841-5813}
\affiliation{%
  \institution{Shanghai Jiao Tong University}
  \city{Shanghai}
  \country{China}
}
\affiliation{%
  \institution{Microsoft Research Asia}
  \city{Beijing}
  \country{China}
}

\author{Shizhao Sun}
\correspondingauthor
\email{shizhaosun@outlook.com}
\orcid{0009-0009-0172-9729}
\affiliation{%
  \institution{Microsoft Research Asia}
  \city{Beijing}
  \country{China}
}

\author{Meisheng Zhang}
\email{2401210766@stu.pku.edu.cn}
\orcid{0009-0004-9511-2052}
\affiliation{%
  \institution{Microsoft Research Asia}
  \city{Beijing}
  \country{China}
}
\affiliation{%
  \institution{Peking University}
  \city{Beijing}
  \country{China}
}

\author{Yingdong Shi}
\email{shiyd2023@shanghaitech.edu.cn}
\orcid{0009-0003-9346-804X}
\affiliation{%
  \institution{Microsoft Research Asia}
  \city{Beijing}
  \country{China}
}
\affiliation{%
  \institution{ShanghaiTech University}
  \city{Shanghai}
  \country{China}
}

\author{Xubo Yang}
\correspondingauthor
\email{yangxubo@sjtu.edu.cn}
\orcid{0000-0001-5378-4003}
\affiliation{%
  \institution{Shanghai Jiao Tong University}
  \city{Shanghai}
  \country{China}
}

\author{Jiang Bian}
\email{jiang.bian.prc@gmail.com}
\orcid{0000-0002-9472-600X}
\affiliation{%
  \institution{Microsoft Research Asia}
  \city{Beijing}
  \country{China}
}

\renewcommand{\shortauthors}{Yang Zhao et al.}

\begin{abstract}
Current one-pass 3D scene synthesis methods often suffer from spatial hallucinations, such as collisions, due to a lack of deliberative reasoning. To bridge this gap, we introduce SceneReVis, a vision-grounded self-reflection framework that employs an iterative ``diagnose-and-act'' loop to explicitly intercept and resolve spatial conflicts using multi-modal feedback. To support this step-wise paradigm, we construct SceneChain-12k, a large-scale dataset of causal construction trajectories derived through a novel reverse engineering pipeline. We further propose a two-stage training recipe that transitions from Supervised Fine-Tuning to Agentic Reinforcement Learning, evolving the model into an active spatial planner. Extensive experiments demonstrate that SceneReVis achieves state-of-the-art performance in high-fidelity generation and goal-oriented optimization, with improved transfer to underrepresented room categories in the dataset and promising potential for extension to out-of-distribution room types.
\end{abstract}



\maketitle

\begin{figure*}[!t]
  \centering
  \includegraphics[width=\textwidth]{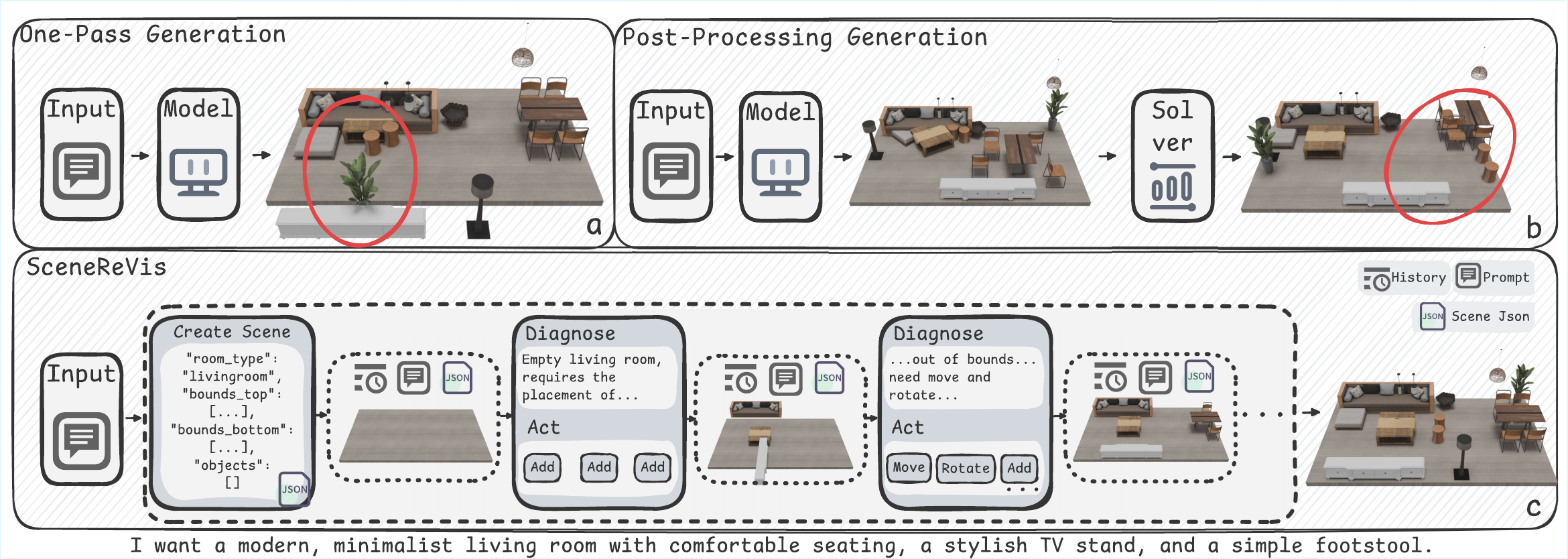}
  \caption{\textbf{Comparison of generation paradigms.}
  \textbf{(a) One-Pass Generation} lacks intermediate reasoning, leading to severe physical violations.
  \textbf{(b) Post-Processing Generation} tends to be trapped in local optima, yielding suboptimal visual and semantic quality.
  \textbf{(c) SceneReVis (Ours)} employs an self-reflection paradigm to ensure physical plausibility and aesthetic coherence.}
  \Description{Comparison of one-pass, post-processing, and SceneReVis generation paradigms with corresponding scene quality outcomes.}
  \label{fig:teaser}
\end{figure*}

\section{Introduction}

Scalable and diverse 3D environments are critical infrastructures for the era of spatial computing, underpinning advancements in Embodied AI, immersive gaming, and automated content creation. For these virtual worlds to be actionable, they demand strict adherence to both semantic coherence and physical laws. In the research community, 3D indoor scene synthesis is typically formulated as a layout prediction task, determining the arrangement of objects represented by 3D bounding boxes and semantic labels~\cite{fu20213dfront,paschalidou2021atiss}. However, automating this process to generate high-fidelity scenes that rival manual design remains a formidable challenge, largely due to the complexity of satisfying intricate geometric and functional constraints. 

Current paradigms~\cite{feng2023layoutgpt,paschalidou2021atiss,tang2024diffuscene} primarily treat scene generation as a static, ``one-pass" prediction task—directly mapping input prompts to a complete layout configuration in a single inference step. Analogous to ``System 1" thinking, this approach relies on rapid, intuitive pattern matching. However, it fundamentally lacks the ``System 2" deliberative reasoning required for iterative self-correction~\cite{evans1984heuristic,kahneman2003maps}, inevitably resulting in ``Spatial Hallucinations"—such as floating objects and severe collisions. Although some approaches~\cite{yang2024holodeck,ccelen2024design} attempt to mitigate these errors via constraint-based post-processing, they heavily rely on the quality of the initial scene priors. Once the initialization exhibits significant structural flaws—such as extreme overcrowding or sparsity—this paradigm effectively becomes trapped in local optima, making it incapable of correcting fundamental layout deficiencies.

To bridge the gap between intuition and reasoning, we introduce \textbf{SceneReVis}, an on-the-fly self-reflection framework that transforms scene synthesis into a dynamic, step-wise decision process. SceneReVis operates through an iterative ``diagnose-and-act" loop(Figure~\ref{fig:teaser}(c)). At each timestep, it perceives the evolving environment via multi-modal feedback—synergizing rendered visual observations with semantic scene graphs. Based on this real-time diagnosis, it explicitly selects and executes atomic operations from a discrete toolset to progressively construct or adjust the layout. Crucially, this mechanism fundamentally diverges from one-pass generation and traditional post-processing strategies. Instead of attempting to blindly output a complete layout at once or waiting to repair a fully generated but flawed scene, SceneReVis performs early error interception: it detects and resolves issues the moment they emerge. This ``fix-as-you-go'' mechanism is critical for two reasons: physically, it maintains a valid geometric foundation at every step, avoiding the entangled spatial conflicts typical of one-pass predictions which are difficult to decouple post-hoc; semantically and aesthetically, the continuous visual feedback enables SceneReVis to actively align objects based on visual harmony.

The first fundamental challenge in training such a multi-turn framework is the scarcity of process-oriented data. Existing 3D datasets~\cite{fu20213dfront,zhonginternscenes,yu2025metascenes,khanna2024habitat} offer only static final snapshots, lacking the temporal trajectories required to learn step-wise generation. In this work, we choose 3D-FRONT as the source corpus because it remains the most mainstream and widely adopted benchmark in indoor scene synthesis, with broad annotations and standardized evaluation protocols that support fair and reproducible comparison, despite its relatively compact category taxonomy. To address this, we implement a Reverse Engineering Pipeline inspired by EditRoom~\cite{zhengeditroom}. Instead of merely dismantling scenes, we systematically deconstruct high-quality static layouts step-by-step into an empty state and record the reverse construction sequences required to recover the optimal layout. This process generates executable tool invocation sequences, mapping the trajectory from an empty room to the final scene, culminating in our large-scale dataset, \textbf{SceneChain-12k}.

With SceneChain-12k available, a straightforward approach is to apply Supervised Fine-Tuning (SFT) so that the LLM imitates tool-use trajectories in the dataset. However, relying on this paradigm alone still yields unsatisfactory performance: because the SFT objective primarily fits demonstrated trajectories, the model often struggles to stably develop the spatial diagnostic capability and long-horizon planning needed to resolve unseen geometric conflicts in complex multi-step decision processes. This limitation becomes more evident when supervision is sparse; and in real-world scene data, even high-quality and broad-coverage datasets inevitably contain low-frequency and long-tail categories, so such sparse-supervision cases persist. Motivated by this, we advance the training paradigm from SFT to multi-turn Agentic Reinforcement Learning (RL). Utilizing Group Relative Policy Optimization (GRPO)~\cite{guo2025deepseek}, we incorporate rendered visual feedback and apply dense rewards tailored to different turn characteristics, enabling the model to evolve from passively imitating tool calls into an active spatial planning policy that continuously adapts to environmental feedback, thereby better optimizing non-differentiable objectives such as physical rigor and aesthetics.

In summary, our contributions are three-fold: 
\begin{itemize}
    \item We formulate 3D scene generation as a vision-grounded sequential decision-making process. To address this, we propose \textbf{SceneReVis}, a vision-grounded self-reflection framework that utilizes visual feedback to proactively intercept and resolve spatial conflicts.
    \item We introduce a holistic pipeline utilizing Reverse Engineering to synthesize the process-oriented dataset, \textbf{SceneChain-12k}. We further employ a two-stage training recipe—advancing from SFT to Agentic RL—to evolve the agent into an active spatial planner.
    \item We demonstrate that \textbf{SceneReVis} achieves state-of-the-art performance in text-conditioned standard scene generation, generalizes strongly to underrepresented (low-frequency) room categories in the dataset, shows potential in out-of-distribution (OOD) scenarios, and supports goal-oriented scene optimization.
\end{itemize}

\begin{table*}[t]
\centering
\caption{\textbf{High-level capability comparison in text-conditioned 3D indoor scene synthesis.}}
\label{tab:intro_capability_comparison}
\setlength{\tabcolsep}{4.5pt}
\resizebox{\textwidth}{!}{%
\begin{tabular}{llccccccc}
\toprule
\textbf{Paradigm} & \textbf{Method} & \textbf{Input} & \textbf{Data-driven} & \textbf{Self-reflection} & \textbf{Visual Feedback} & \textbf{Collision Detection} & \textbf{Irregular Floor Support} & \textbf{Goal-oriented Optimization} \\
\midrule
\multirow{5}{*}{One-Pass} 
 & LayoutGPT~\cite{feng2023layoutgpt} & Text & \xmark & \xmark & \xmark & \xmark & \xmark & \xmark \\
 & DiffuScene~\cite{tang2024diffuscene} & Text / Scene & \cmark & \xmark & \xmark & \xmark & \xmark & \xmark \\
 & PhyScene~\cite{yang2024physcene} & Floor Plan & \cmark & \xmark & \xmark & \xmark & \cmark & \xmark \\
 & ReSpace~\cite{bucher2025respace} & Text (+ Scene) & \cmark & \xmark & \xmark & \xmark & \cmark & \xmark \\
 & OptiScene~\cite{yang2025optiscene} & Text & \cmark & \xmark & \xmark & \xmark & \xmark & \xmark \\
\midrule
\multirow{5}{*}{Post-Processing} 
 & Holodeck~\cite{yang2024holodeck} & Text & \xmark & \xmark & \xmark & \cmark & \xmark & \xmark \\
 & I-Design~\cite{ccelen2024design} & Text & \xmark & \xmark & \xmark & \cmark & \xmark & \xmark \\
 & LayoutVLM~\cite{sun2025layoutvlm} & Text + Scene & \xmark & \xmark & \cmark & \cmark & \xmark & \xmark \\
 & DirectLayout~\cite{ran2025direct} & Text & \cmark & \xmark & \cmark & \xmark & \xmark & \xmark \\
 & SceneThesis~\cite{ling2025scenethesis} & Text & \xmark & \xmark & \cmark & \cmark & \xmark & \xmark \\
\midrule
\multirow{4}{*}{Agentic} 
 & SceneWeaver~\cite{yangsceneweaver} & Text & \xmark & \cmark & \cmark & \cmark & \xmark & \xmark \\
 & SAGE~\cite{xia2026sage} & Text (+ Image) & \xmark & \cmark & \cmark & \cmark & \xmark & \xmark \\
 & SceneSmith~\cite{scenesmith2026} & Text & \xmark & \cmark & \cmark & \cmark & \xmark & \xmark \\
\cmidrule{2-9}
 & \textbf{SceneReVis (Ours)} & \textbf{Text (+ Scene)} & \textbf{\cmark} & \textbf{\cmark} & \textbf{\cmark} & \textbf{\cmark} & \textbf{\cmark} & \textbf{\cmark} \\
\bottomrule
\end{tabular}%
}
\end{table*}
\section{Related work}

\textbf{3D Indoor Scene Synthesis.} Prior work on indoor scene synthesis can be organized into three paradigms (Table~\ref{tab:intro_capability_comparison}). \textbf{(1) One-Pass Generation}~\cite{bucher2025respace,feng2023layoutgpt,yang2025optiscene,paschalidou2021atiss,yang2024physcene,pan2025metaspatial,tang2024diffuscene,lin2024instructscene,choi2026scenenat} predicts full layouts from learned priors (e.g., 3D-FRONT~\cite{fu20213dfront} and language models). This design is efficient, but explicit mechanisms for correcting initial predictions are limited, so physical conflicts may persist. \textbf{(2) Post-Processing}~\cite{ccelen2024design,yang2024holodeck,sun2025layoutvlm,ran2025direct,ling2025scenethesis,gao2025disco} includes two common subtypes: solver-based refinement to improve feasibility, and regenerate-and-refine pipelines (e.g., DirectLayout~\cite{ran2025direct} and SceneThesis~\cite{ling2025scenethesis}) that re-generate layouts when requirements are not met. The former can prioritize constraint satisfaction over fine visual quality, while the latter can increase runtime due to repeated generation. \textbf{(3) Agentic Generation}~\cite{yangsceneweaver,scenesmith2026,xia2026sage} introduces iterative feedback, with concurrent methods such as SceneSmith~\cite{scenesmith2026} and SAGE~\cite{xia2026sage} further incorporating object-level physical attributes. Many existing pipelines still rely on frozen LLMs plus external tool orchestration; without task-specific tuning, pure next-token prediction may emphasize linguistic plausibility more than spatial correctness. In contrast, \textbf{SceneReVis} uses Agentic RL to internalize spatial priors into the policy, increasing probability mass on physically feasible actions.

\textbf{Spatial Chain-of-Thought.} Recent advancements have applied Chain-of-Thought (CoT) reasoning to 3D synthesis tasks. However, existing approaches~\cite{ran2025direct,haomesatask} predominately function as open-loop text planners, deriving layouts solely from linguistic priors without access to rendered visual feedback. Consequently, they often suffer from spatial hallucinations and lack the capability to self-correct during generation. In contrast, we propose a Vision-Grounded Spatial Diagnostic CoT that integrates visual perception into the reasoning loop by learning from editing chains constructed via reverse engineering, thereby enabling the model to explicitly master dynamic ``diagnose-and-act'' capabilities for precise spatial adjustment.

\textbf{Reinforcement Learning in 3D Scene Synthesis.} Compared to prior RL exploration in 3D synthesis, existing approaches~\cite{bucher2025respace,haomesatask,ran2025direct,yang2025optiscene,pan2025metaspatial} primarily leverage DPO~\cite{rafailov2023direct} or single-turn RL for alignment. However, DPO is hindered by restricted exploration due to its reliance on static offline data, while single-turn RL operates as a one-pass mechanism, lacking the interactive feedback loop required to self-correct intermediate errors. In contrast, we formulate synthesis as a multi-turn POMDP to enable iterative refinement.
Furthermore, while general Agentic RL has succeeded in text-based reasoning~\cite{wang2025practitioner,feng2025retool,singh2025agentic,dong2025agentic}, applying such paradigms to 3D presents unique challenges, particularly in grounding abstract reasoning into physical reality. To bridge this gap, we integrate visual and physical grounding directly into the optimization loop, empowering the agent to align spatial reasoning with physical reality beyond the reach of pure linguistic priors.

\begin{figure*}[!t]
   \centering
   \includegraphics[width=\textwidth]{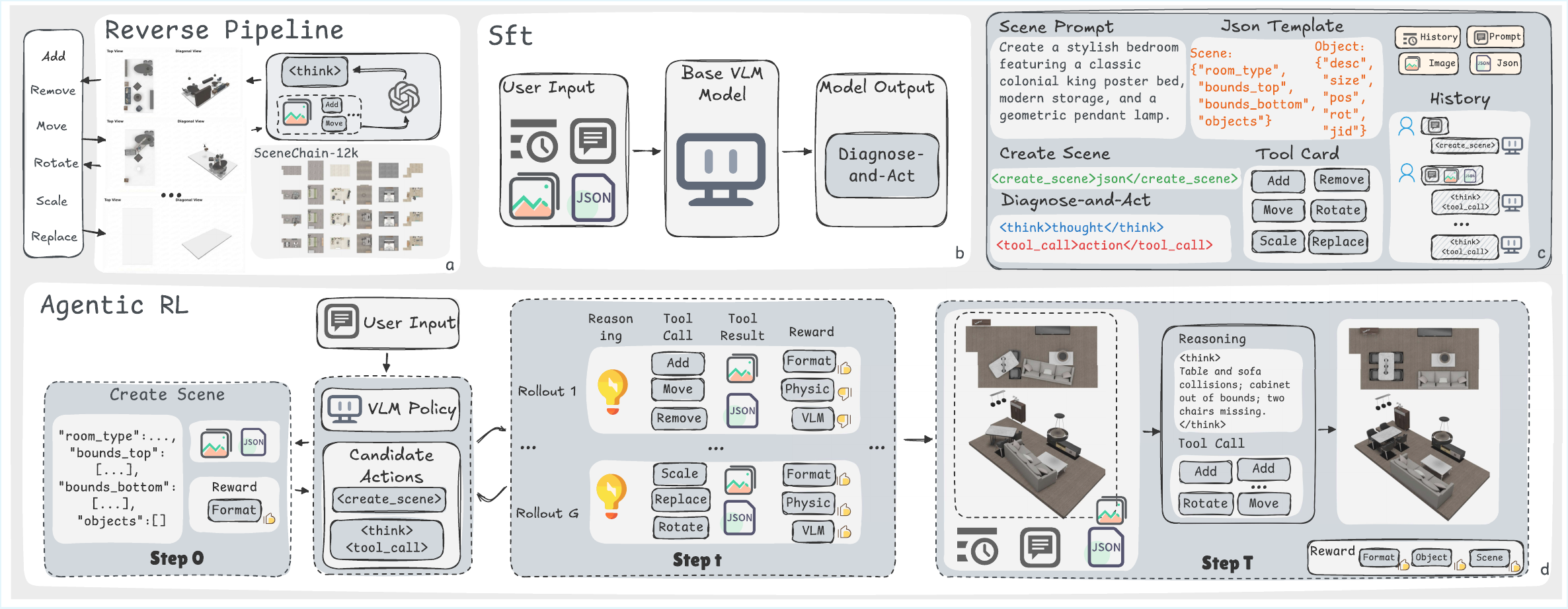}
   \caption{Overview of the SceneReVis learning framework. The pipeline consists of three stages: (a) \textbf{Data Construction:} We employ a Reverse Engineering strategy to decompose static 3D scenes into causal construction trajectories (SceneChain-12k). (b) \textbf{Cold Start:} A Supervised Fine-Tuning (SFT) stage initializes the agent with basic tool usage capabilities. (c) \textbf{Legend:} Legend explaining the contents of various parts. (d) \textbf{Agentic RL:} The core stage utilizes Group Relative Policy Optimization (GRPO). The agent interacts with a physics-enabled simulator via a "diagnose-and-act" loop, receiving multi-modal feedback and optimizing against non-differentiable objectives.}
   \Description{Training pipeline of SceneReVis with data construction, SFT cold start, and agentic RL optimization.}
   \label{fig:pipeline}
\end{figure*}
\section{Method}

We formulate the task as a POMDP defined by the tuple $\mathcal{M} = (\mathcal{S}, \mathcal{A}, \mathcal{T}, \mathcal{R}, \Omega, \mathcal{O}, \gamma)$. Here, $\mathcal{S}$ denotes the latent simulator state, projected by $\mathcal{O}$ into the observation space $\Omega$. As visualized in the Figure~\ref{fig:pipeline}(c), the observation $o_t = (I, G_t, V_t, H_{t-1}) \in \Omega$ integrates the instruction $I$, structured scene graph $G_t$, visual rendering $V_t$, and interaction history $H_{t-1}$. The agent generates composite actions $a_t \in \mathcal{A}$ to update the scene via the transition function $\mathcal{T}$. The objective is to maximize the return defined by reward $\mathcal{R}$ and discount $\gamma$. In all experiments, we set $\gamma = 1$.

The remainder of this section details the five core components of our framework: (1) a vision-grounded iterative reasoning mechanism, (2) two-stage policy alignment, (3) automated data synthesis via reverse engineering, (4) a hybrid reward formulation, and (5) extended applications of our framework.

\subsection{Vision-Grounded Self-Reflection Framework}
\label{sec:framework}

The inference mechanism operates as a closed-loop POMDP cycle. The process initiates by instantiating an initial scene graph based on the user instruction to form the initial state $s_0$. At each step $t$, the policy $\pi_\theta$ perceives a multi-modal observation $o_t = (I, G_t, V_t, H_{t-1})$, integrating the global instruction, current semantic and visual states, and the cumulative interaction history. The rendered feedback $V_t$ is particularly critical for exposing fine-grained geometric details latent in the structured graph.

Adopting a \textit{Diagnose-and-Act} paradigm, SceneReVis first generates a Spatial Diagnosis CoT to identify semantic gaps or physical violations, yielding a structured tool invocation $u_t \in \mathcal{A}_{\text{prim}}$. This primitive space $\mathcal{A}_{\text{prim}}$ encompasses atomic actions for instantiation (\texttt{Add}), affine transformations (\texttt{Move}, \texttt{Rotate}, \texttt{Scale}), and semantic updates (\texttt{Replace}, \texttt{Remove}).

Driven by this decision, the environment transition $\mathcal{T}$ executes a retrieval-augmented pipeline to realize the instantiation primitives specified in $u_t$. Specifically, we employ a coarse-to-fine strategy: a candidate set is first retrieved based on the semantic category predicted in $u_t$, followed by selecting the asset that minimizes the aspect ratio alignment error with the target dimensions generated by the agent.

Inspired by~\cite{ccelen2024design,yang2024holodeck,yangsceneweaver}, we mitigate residual collisions via a lightweight optimization step that applies pose micro-perturbations and prunes overlapping entities to ensure physical plausibility in $\mathcal{S}$.

\subsection{Two-Stage Policy Learning}

Our training framework consists of two sequential phases: supervised initialization to establish policy priors, and reinforcement learning for objective alignment.

\textbf{Stage I: Structural Priors via SFT (Figure~\ref{fig:pipeline}(b))}
We initialize the policy by fine-tuning a pre-trained MLLM on flattened trajectories formatted as multi-turn dialogues. In each turn, the model processes the observation $o_t = (I, G_t, V_t, H_{t-1})$ to predict a composite response $a_t = (z_t, u_t)$, comprising the spatial diagnostic CoT and executable tool call. Treating this as a standard sequence modeling task, we maximize the likelihood of the agent's action tokens conditioned on the interaction history. We apply a loss masking strategy to ensure gradients are backpropagated exclusively on agent outputs, ignoring user instructions and visual tokens. This cold-start phase serves a dual purpose: (1) \textit{Syntactic Alignment}, forcing strict adherence to the output protocol; and (2) \textit{Semantic Prior Injection}, allowing the model to implicitly clone the spatial layout priors embedded in the static demonstrations. To achieve it, we construct a dataset of multi-turn CoT trajectories, the synthesis details of which are elaborated in the Section~\ref{data_synthesis}.

\textbf{Stage II: Objective Alignment via GRPO (Figure~\ref{fig:pipeline}(d))}
To enable the ``diagnose-and-act'' capability, we formulate the optimization as a multi-stage sequential decision process driven by visual feedback. Utilizing Group Relative Policy Optimization (GRPO) as the backbone, we structure the learning curriculum into three distinct phases. At the initial step ($t=0$), the agent establishes the global context by defining the scene's template, shape, and dimensions. During intermediate turns ($0 < t < T$), the agent executes the iterative ``diagnose-and-act'' loop to populate objects and rectify spatial conflicts. Crucially, each turn incorporates visual feedback from rendered top-down and random diagonal views, allowing the agent to perceive geometric relations explicitly. Finally, at the terminal step ($t=T$), the completed layout undergoes a comprehensive quality assessment. This structured progression enables us to apply targeted supervision across different stages, serving as the foundation for the reward design detailed in Section~\ref{method_reward}.

\subsection{Synthesis of Multi-Turn Reasoning Trajectories}
\label{data_synthesis}
To equip the agent with long-horizon planning capabilities, we construct \textbf{SceneChain-12k}(Figure~\ref{fig:pipeline}(a)), a large-scale instruction-tuning dataset $\mathcal{D} = \{(I, \tau)\}$. For each scene, the trajectory is formulated as $\tau = (o_0, a_0, \dots, o_T)$, where the composite action $a_t = (z_t, u_t)$ consists of a \textit{Spatial Diagnostic Chain-of-Thought (CoT)} $z_t$ and an executable tool call $u_t$. The construction of $\mathcal{D}$ follows a rigorous three-stage pipeline:

\textbf{Step 1: Primitive Sequence Generation via Reverse Engineering.}
First,  we generate the ground-truth primitive action sequence $u_{0:T-1}$ of length $T$ by recursively dismantling high-fidelity scenes. Utilizing 3D-FRONT~\cite{fu20213dfront} as the raw source, we filter 4,059 valid scenes and employ a \textit{Scene Reverse Editor} to stochastically sample inverse primitives $\hat{u}_k$ from the atomic space $\mathcal{A}_{\text{prim}}$ (\texttt{Add}, \texttt{Remove}, \texttt{Move}, $\ldots$, \texttt{Replace}).~\cite{zhengeditroom} The recorded deconstruction chain is then temporally inverted ($u_t = \hat{u}^{-1}_{T-1-t}$) to form the forward trajectory. Crucially, to align with human design priors, we implement a \textit{Stage-Aware Deconstruction} strategy: the sampling probability is conditioned on object volume relative to the editing progress. Small decorative items are prioritized for removal in early reverse stages, while large structural entities are retained until the end, mathematically enforcing a logical ``coarse-to-fine'' hierarchy in the final forward generation.

\textbf{Step 2: Quality Assurance via Automated Filtration.}
Second, we screen the generated sequences to eliminate suboptimal behaviors derived from stochastic inversion. We synthesize $N=10$ candidate trajectories for each scene and employ GPT-5 to evaluate them using a multi-dimensional scoring system. Focusing on \textit{Editing Coherence}, \textit{Naturalness}, and \textit{Transition Quality}, we retain only the top-3 trajectories that satisfy strict physical constraints. This rigorous selection process yields a final corpus of \textbf{12,177} high-quality samples, effectively purging invalid or unnatural transitions.

\textbf{Step 3: Diagnostic CoT Synthesis.}
Finally, we synthesize the diagnostic reasoning $z_t$ to bridge the visual state $s_t$ and the tool call $u_t$. We reframe scene construction as a sequential optimization process. Conditioned on the visual state, we prompt the LLM to generate reasoning following a structured taxonomy of \textit{Scene Bugs}: (1) \textit{Physical Conflicts}, (2) \textit{Layout Rationality}, and (3) \textit{Spatial Distribution}. The synthesized CoT enforces a strict logic flow: Spatial Diagnosis (identifying the bug) $\to$ Patching Strategy (formulating a fix) $\to$ Execution (tool invocation), transforming the generation task into a reasoned constraint-satisfaction problem.

\subsection{Policy Optimization via Hybrid Feedback Mechanisms}
\label{method_reward}
Training an agent to generate complex 3D scenes requires balancing structural validity, physical feasibility, and semantic alignment. To guide the policy $\pi_\theta$ efficiently, we construct a composite reward mechanism operating across three temporal phases.

\textbf{Structural Initialization ($t=0$).} 
The process initiates by constructing the geometric scaffold. At this stage, the objective is to ensure the room layout is well-formed. We define an initialization reward $r_{init}$ to enforce Schema Compliance and Geometric Validity, strictly penalizing fatal errors such as unparsable JSON outputs, as well as degenerate room geometries (e.g., overly irregular shapes or dimensions violating realistic size constraints), ensuring the episode starts from a valid state.

\textbf{Iterative Refinement ($0 < t < T$).} 
During the generation process, we provide dense feedback to steer the agent. The intermediate reward $r_t$ is computed as a weighted sum:
\begin{equation}
\begin{split}
    r_t(s_t, a_t) = & \, \lambda_{fmt} r_{fmt}(a_t) + \lambda_{phy} r_{phy}(s_t) \\
    & + \lambda_{sem} r_{sem}(s_t, s_{t-1})
\end{split}
\end{equation}
The format term $r_{fmt}$ ensures the generated tool calls follow valid syntax for execution. The physical term $r_{phy}$ penalizes immediate collisions and out-of-bounds errors detected by the physics engine. Crucially, the semantic term $r_{sem}$ serves as a Goal Progress Signal. Instead of just assessing the current state, it evaluates the \textit{semantic delta} between $s_t$ and $s_{t-1}$ (e.g., successfully adding a required object from the instruction), explicitly penalizing stagnation or degradation to encourage efficient, step-wise planning.

\textbf{Terminal Assessment and Trajectory Aggregation.} 
At the conclusion ($t=T$), we calculate a Hierarchical Final Reward $R_{final}$ to assess the holistic quality:
\begin{equation}
\begin{split}
    R_{final}(s_T) = & \, \omega_{fmt} \mathcal{R}_{fmt}(s_T) + \omega_{obj} \mathcal{R}_{obj}(s_T) \\
    & + \omega_{scene} \mathcal{R}_{scene}(s_T)
\end{split}
\end{equation}
where components are normalized to $[-1, 1]$. $\mathcal{R}_{fmt}$ verifies final output structure. $\mathcal{R}_{obj}$ checks the completeness of core furniture defined in the instruction and validates the correctness of object scales. $\mathcal{R}_{scene}$ combines rigid physical rules to judge overall physical plausibility with a VLM-based evaluator to score visual aesthetics and semantic coherence. Finally, the total trajectory reward $J(\tau)$ aggregates the mean intermediate reward and the final reward to balance safe execution with high-fidelity results:
\begin{equation}
    J(\tau) =  \, \alpha \cdot \underbrace{\left( \frac{1}{T} \sum_{t=0}^{T-1} r_t(s_t, a_t) \right)}_{\text{Process Consistency}} + \beta \cdot \underbrace{R_{final}(s_T)}_{\text{Outcome Quality}}
\end{equation}
This formulation ensures the agent is penalized for intermediate invalid actions while primarily incentivizing the semantic and physical integrity of the final layout. This undiscounted aggregation is consistent with our POMDP setting $\gamma=1$.

\begin{table*}[htbp]
\centering
\caption{\textbf{Quantitative comparison on Bedroom and Living Room.} `Avg.` is the mean of Ra, Spa, and Ac. Best results are \textbf{bolded}, and second-best results are \underline{underlined}.}
\label{tab:comparison_bed_living}

\setlength{\tabcolsep}{3.5pt}

\resizebox{\textwidth}{!}{%
\begin{tabular}{l cccccc cccccc cccccc}
\toprule

\multirow{3}{*}{\textbf{Method}} & \multicolumn{6}{c}{\textbf{Bedroom}} & \multicolumn{6}{c}{\textbf{Living Room}} & \multicolumn{6}{c}{\textbf{Avg. (Bed + Living)}} \\
\cmidrule(lr){2-7} \cmidrule(lr){8-13} \cmidrule(lr){14-19}
 & \multicolumn{2}{c}{Physics} & \multicolumn{4}{c}{Visual \& Semantics} 
 & \multicolumn{2}{c}{Physics} & \multicolumn{4}{c}{Visual \& Semantics}
 & \multicolumn{2}{c}{Physics} & \multicolumn{4}{c}{Visual \& Semantics} \\
\cmidrule(lr){2-3} \cmidrule(lr){4-7} \cmidrule(lr){8-9} \cmidrule(lr){10-13} \cmidrule(lr){14-15} \cmidrule(lr){16-19}
 & OBR$\downarrow$ & CNR$\downarrow$ & Ra.$\uparrow$ & Spa.$\uparrow$ & Ac.$\uparrow$ & \textbf{Avg.}$\uparrow$
 & OBR$\downarrow$ & CNR$\downarrow$ & Ra.$\uparrow$ & Spa.$\uparrow$ & Ac.$\uparrow$ & \textbf{Avg.}$\uparrow$
 & OBR$\downarrow$ & CNR$\downarrow$ & Ra.$\uparrow$ & Spa.$\uparrow$ & Ac.$\uparrow$ & \textbf{Avg.}$\uparrow$ \\
\midrule

DiffuScene~\cite{tang2024diffuscene} & 46.2 & 33.8 & 8.2 & 6.2 & 6.9 & 7.1 & 28.0 & 41.5 & 7.3 & 7.5 & 8.2 & 7.7 & 37.1 & 37.7 & 7.8 & 6.9 & 7.6 & 7.4 \\

ReSpace~\cite{bucher2025respace}    & 14.7 & 41.9 & 6.8 & 6.1 & 7.0 & 6.6 & 11.5 & 40.3 & 6.9 & 6.6 & 7.3 & 6.9 & 13.1 & 41.1 & 6.9 & 6.4 & 7.2 & 6.8 \\
\cmidrule(lr){1-19} 
LayoutGPT~\cite{feng2023layoutgpt}  & 34.3 & 45.9 & 7.8 & 5.6 & 7.6 & 7.0 & 15.0 & 35.7 & 7.5 & 7.3 & \underline{8.7} & \underline{7.8} & 24.7 & 40.8 & 7.7 & 6.5 & 8.2 & 7.5 \\

I-Design~\cite{ccelen2024design}   & 15.5 & 18.2 & 8.3 & 6.6 & 8.1 & \underline{7.7} & 17.8 & 14.1 & \underline{8.3} & 6.5 & 8.6 & \underline{7.8} & 16.7 & 16.2 & \underline{8.3} & 6.6 & 8.4 & \underline{7.8} \\

Holodeck~\cite{yang2024holodeck}   & 15.4 & 22.2 & 7.1 & \textbf{8.1} & \underline{8.6} & \underline{7.9} & 10.0 & \textbf{3.1} & 5.9 & \textbf{8.2} & \textbf{8.8} & 7.6 & 12.7 & 12.7 & 6.5 & \textbf{8.2} & \underline{8.7} & \underline{7.8} \\

LayoutVLM~\cite{sun2025layoutvlm}  & 11.5 & 44.9 & 7.0 & 7.6 & 8.0 & 7.5 & 14.3 & 28.7 & 6.8 & \underline{7.9} & \underline{8.7} & \underline{7.8} & 12.9 & 36.8 & 6.9 & 7.8 & 8.4 & 7.7 \\
SceneWeaver~\cite{yangsceneweaver} & \textbf{0.0} & \underline{11.0} & \underline{8.4} & 7.3 & 7.4 & 7.7 & \textbf{0.7} & 8.0 & 7.2 & 7.1 & 8.4 & 7.6 & \textbf{0.4} & \underline{9.5} & 7.8 & 7.2 & 7.9 & 7.6 \\
\cmidrule(lr){1-19} 
\textbf{Ours} & \underline{2.8} & \textbf{4.6} & \textbf{9.3} & \underline{8.0} & \textbf{8.9} & \textbf{8.7} & \underline{1.2} & \underline{4.4} & \textbf{9.5} & \underline{8.0} & \textbf{8.8} & \textbf{8.8} & \underline{2.0} & \textbf{4.5} & \textbf{9.4} & \underline{8.0} & \textbf{8.9} & \textbf{8.8} \\

\bottomrule
\end{tabular}%
}
\end{table*}

\begin{table*}[htbp]
\centering
\caption{\textbf{Quantitative comparison on Dining Room and Study Room.} `Avg.` is the mean of Ra, Spa, and Ac. Best results are \textbf{bolded}, and second-best results are \underline{underlined}.}
\label{tab:comparison_dining_study_avg}

\setlength{\tabcolsep}{3.5pt}

\resizebox{\textwidth}{!}{%
\begin{tabular}{l cccccc cccccc cccccc}
\toprule

\multirow{3}{*}{\textbf{Method}} & \multicolumn{6}{c}{\textbf{Dining Room}} & \multicolumn{6}{c}{\textbf{Study Room}} & \multicolumn{6}{c}{\textbf{Avg. (Dining + Study)}} \\
\cmidrule(lr){2-7} \cmidrule(lr){8-13} \cmidrule(lr){14-19}
 & \multicolumn{2}{c}{Physics} & \multicolumn{4}{c}{Visual \& Semantics} 
 & \multicolumn{2}{c}{Physics} & \multicolumn{4}{c}{Visual \& Semantics}
 & \multicolumn{2}{c}{Physics} & \multicolumn{4}{c}{Visual \& Semantics} \\
\cmidrule(lr){2-3} \cmidrule(lr){4-7} \cmidrule(lr){8-9} \cmidrule(lr){10-13} \cmidrule(lr){14-15} \cmidrule(lr){16-19}
 & OBR$\downarrow$ & CNR$\downarrow$ & Ra.$\uparrow$ & Spa.$\uparrow$ & Ac.$\uparrow$ & \textbf{Avg.}$\uparrow$
 & OBR$\downarrow$ & CNR$\downarrow$ & Ra.$\uparrow$ & Spa.$\uparrow$ & Ac.$\uparrow$ & \textbf{Avg.}$\uparrow$
 & OBR$\downarrow$ & CNR$\downarrow$ & Ra.$\uparrow$ & Spa.$\uparrow$ & Ac.$\uparrow$ & \textbf{Avg.}$\uparrow$ \\
\midrule

LayoutGPT~\cite{feng2023layoutgpt} & 6.2 & 26.2 & 6.9 & 7.7 & \textbf{8.8} & \underline{7.8} & 16.6 & 52.1 & \textbf{8.5} & 7.2 & 8.3 & \underline{8.0} & 11.4 & 39.2 & 7.7 & 7.5 & 8.6 & \underline{7.9} \\

I-Design~\cite{ccelen2024design}  & 17.6 & \underline{0.7} & \underline{7.8} & 5.7 & \textbf{8.8} & 7.4 & 16.3 & \underline{4.5} & \underline{8.2} & 6.9 & \textbf{8.9} & \underline{8.0} & 17.0 & \underline{2.6} & \underline{8.0} & 6.3 & \textbf{8.9} & 7.7 \\

Holodeck~\cite{yang2024holodeck}  & 11.9 & 3.3 & 6.4 & \underline{8.0} & 8.5 & 7.6 & 13.7 & 5.1 & 6.7 & \textbf{8.4} & \textbf{8.9} & \underline{8.0} & 12.8 & 4.2 & 6.6 & \underline{8.2} & 8.7 & 7.8 \\

LayoutVLM~\cite{sun2025layoutvlm} & 19.2 & 27.0 & 6.5 & 7.8 & \textbf{8.8} & 7.7 & 12.3 & 36.6 & 6.9 & 7.6 & \underline{8.6} & 7.7 & 15.8 & 31.8 & 6.7 & 7.7 & 8.7 & 7.7 \\
SceneWeaver~\cite{yangsceneweaver} & \textbf{0.0} & \textbf{0.3} & 6.3 & 6.6 & 8.3 & 7.1 & \textbf{0.0} & 6.7 & 6.8 & \textbf{8.4} & 8.5 & 7.9 & \textbf{0.0} & 3.0 & 6.5 & 7.5 & 8.4 & 7.5 \\
\cmidrule(lr){1-19} 
\textbf{Ours} & \underline{0.1} & 2.8 & \textbf{8.1} & \textbf{8.3} & \underline{8.6} & \textbf{8.3} & \underline{0.5} & \textbf{2.1} & \underline{8.2} & \underline{8.3} & \textbf{8.9} & \textbf{8.5} & \underline{0.3} & \textbf{2.5} & \textbf{8.2} & \textbf{8.3} & \underline{8.8} & \textbf{8.4} \\

\bottomrule
\end{tabular}%
}
\end{table*}

\subsection{Broad Application}
\label{sec:application}

\textbf{Text-driven Standard Scene Generation.}
This task constructs a complete scene from scratch based on text. Current ``one-pass'' methods~\cite{feng2023layoutgpt,tang2024diffuscene} often suffer from spatial hallucinations, while post-processing approaches~\cite{yang2024holodeck} heavily rely on initialization quality and are prone to getting trapped in local optima. In contrast, our agent builds scenes sequentially from an empty state, leveraging the diagnose-and-act loop to ensure rigorous physical validity without optimization dead-ends.

\textbf{Unified Generalization in Underrepresented and OOD Domains.}
This capability targets two scenarios: generalization to underrepresented room categories within the training distribution (a low-frequency long-tail setting), and exploratory extension to purely out-of-distribution room concepts. Conventional diffusion models~\cite{tang2024diffuscene} struggle here due to fixed vocabularies. Conversely, purely LLM-based methods~\cite{feng2023layoutgpt}, while semantically rich, rely on linguistic priors and lack physical grounding, often yielding chaotic layouts. We bridge this by initializing with a coarse LLM-generated graph and iteratively refining it to enforce physical constraints, effectively decoupling semantic imagination from physical realization.

\textbf{Goal-Oriented Scene Optimization.}
Moving beyond \textit{ab initio} generation, this task refines arbitrary suboptimal layouts—whether chaotic, incomplete, or misaligned—conditioned on an instruction $I$. Unlike prior works~\cite{tang2024diffuscene} that fragment rearrangement and completion into distinct problems, we unify these under a single formulation. Leveraging our \textit{Spatial Diagnostic CoT}, a single policy $\pi_\theta$ iteratively repairs geometric violations and completes missing components, steering any configuration toward physical plausibility.

\begin{figure*}[t!]
  \centering
  \includegraphics[width=1.0\linewidth]{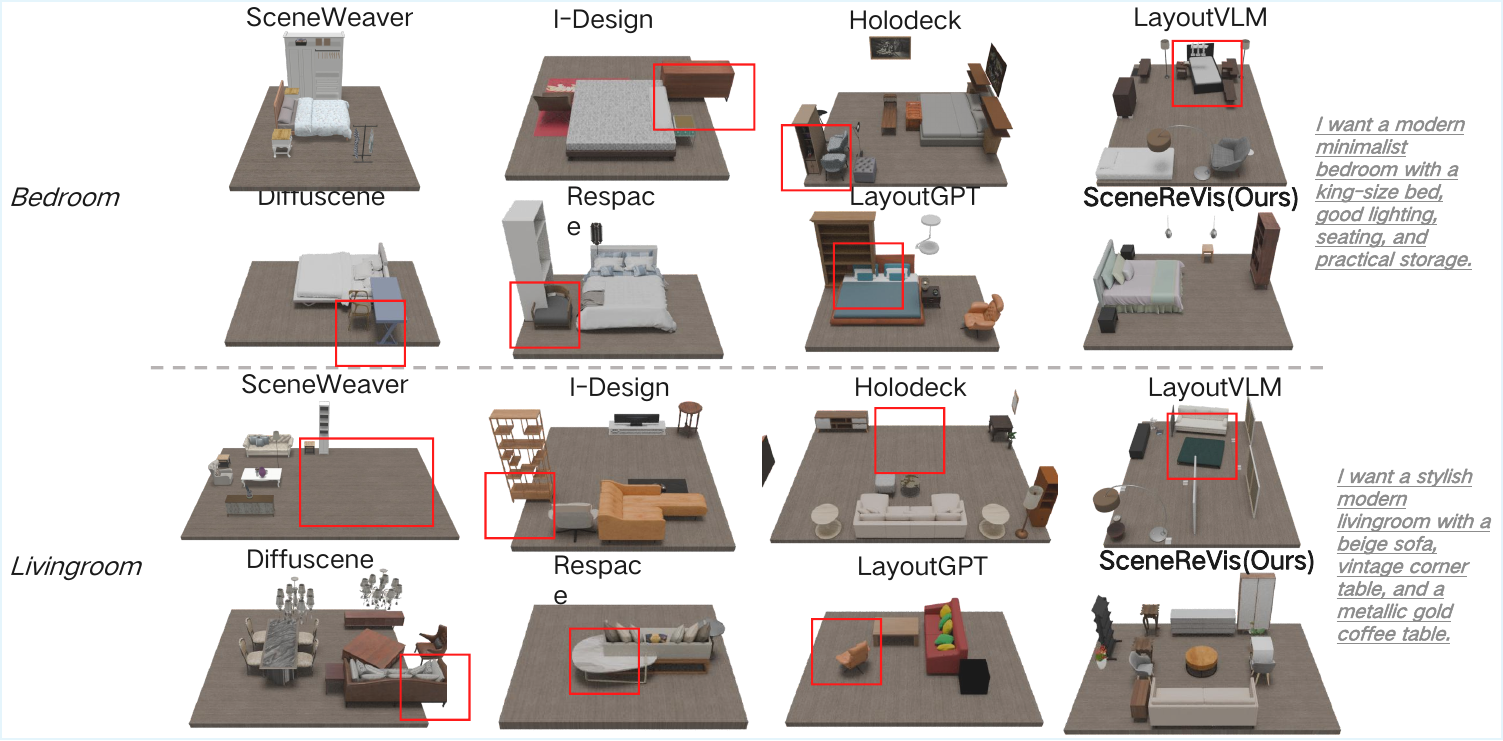} 
  \caption{\textbf{Qualitative comparison of standard scenes.} 
  We visualize the 3D scenes generated by different methods for \textit{Bedroom} (top) and \textit{Living Room} (bottom) scenarios based on the given text prompts.}
  \Description{Qualitative comparison of generated bedroom and living-room scenes across methods.}
  \label{fig:qualitative_standard} 
  
  \includegraphics[width=1.0\linewidth]{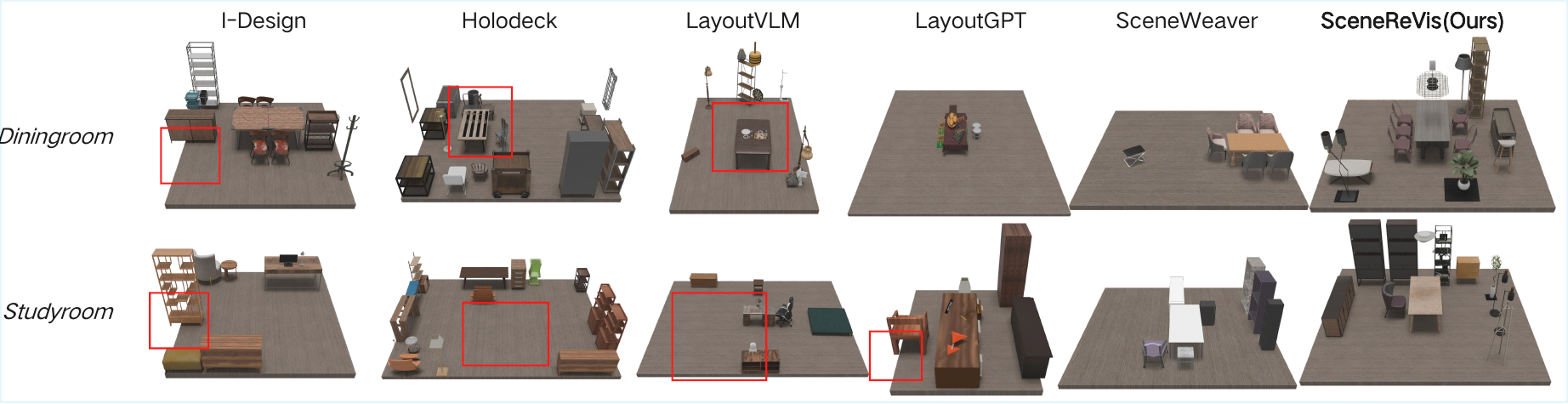} 
  \caption{\textbf{Qualitative comparison on underrepresented room categories.} 
  We visualize the 3D scenes generated by different methods for \textit{Dining room} (top) and \textit{Study Room} (bottom) scenarios based on the given text prompts.}
  \Description{Qualitative comparison of generated dining-room and study-room scenes across methods.}
  \label{fig:qualitative_ood} 
\end{figure*}

\section{Experiment}

\textbf{Datasets.}
We construct \textbf{SceneChain-12k} (12,177 trajectories, 4,059 scenes) from 3D-FRONT~\cite{fu20213dfront}, strictly excluding a held-out pool of 300 trajectories for zero-shot evaluation. \textbf{Training Data} utilizes the remaining trajectories: \textbf{SFT Data} provides instruction-trajectory pairs $(I, \tau)$ (Sec~\ref{data_synthesis}) for behavioral cloning, while \textbf{RL Data} combines 1,000 real training prompts and 1,000 GPT-5 synthesized prompts—introduced to mitigate the scarcity of specific room types—to enable exploration. \textbf{Test Data} comprises two splits: the Standard Split (N=300), which merges 200 real held-out prompts with 100 GPT-synthesized prompts (50 Bedroom, 50 Living Room) for in-distribution testing; and the Generalization Split (N=100), containing synthesized prompts for underrepresented room categories in the dataset (e.g., Dining, Study) to assess generalization in low-frequency settings. See the supplementary material for details.

\textbf{Implementation Details.}
Our policy $\pi_\theta$ is fine-tuned from Qwen2.5-VL-7B-Instruct~\cite{bai2025qwen2}. Following SFT, we continue RL training from the SFT checkpoint using the GRPO algorithm within the VeRL framework~\cite{Shengverl}. Under the NVIDIA B200 configuration, we use Qwen3-VL-235B-A22B-Instruct~\cite{bai2025qwen3vltechnicalreport} as the VLM judge model. Visual observations are rendered via Blender. See the supplementary material for full hyperparameters and hardware specifications.

\textbf{Baselines.}
We benchmark against: (1) Data-driven models (DiffuScene~\cite{tang2024diffuscene}, ReSpace~\cite{bucher2025respace}); (2) LLM-based few-shot learners (LayoutGPT~\cite{feng2023layoutgpt}); and (3) LLM with post-processing methods (I-Design~\cite{ccelen2024design}, Holodeck~\cite{yang2024holodeck}, LayoutVLM~\cite{sun2025layoutvlm}). We also compare against the agentic method SceneWeaver~\cite{yangsceneweaver}. OptiScene~\cite{yang2025optiscene} and DirectLayout\cite{ran2025direct} are omitted due to lack of open-source weights or code. For Goal-oriented Optimization, We specifically compare with DiffuScene and LayoutVLM, two models that are specifically designed for tasks related to scene optimization, under three conditions: \textit{Chaotic \& Missing}, \textit{Chaotic Only}, and \textit{Missing Only}.

\textbf{Metrics.}
We report Physical Fidelity via Out-of-Bounds Rate (OBR) and Collision Rate (CNR), adopting the protocols from~\cite{ran2025direct}. Since physical metrics alone fail to capture high-level functional logic and aesthetics, we introduce Visual \& Semantic Quality assessed by GPT-5 across three dimensions: \textit{Rationality (Ra.)}, \textit{Spatial Layout (Spa.)}, and \textit{Accessibility (Ac.)}. Detailed definitions are provided in the supplementary material.

\begin{table*}[htbp]
\centering
\caption{\textbf{Quantitative comparison on Goal-oriented Scene Optimization tasks.} `Avg.` is the mean of Ra, Spa, and Ac. Best results are \textbf{bolded}, and second-best results are \underline{underlined}.}
\label{tab:goal_oriented_all_in_one}

\setlength{\tabcolsep}{1.5pt}

\resizebox{\textwidth}{!}{%
\begin{tabular}{l cccccc cccccc cccccc cccccc}
\toprule

\multirow{3}{*}{\textbf{Method}} 
 & \multicolumn{6}{c}{\textbf{Cond 1: Chaotic \& Missing}} 
 & \multicolumn{6}{c}{\textbf{Cond 2: Chaotic Only}} 
 & \multicolumn{6}{c}{\textbf{Cond 3: Missing Only}} 
 & \multicolumn{6}{c}{\textbf{Overall Average}} \\
\cmidrule(lr){2-7} \cmidrule(lr){8-13} \cmidrule(lr){14-19} \cmidrule(lr){20-25}

 & \multicolumn{2}{c}{Physics} & \multicolumn{4}{c}{Visual \& Semantics} 
 & \multicolumn{2}{c}{Physics} & \multicolumn{4}{c}{Visual \& Semantics}
 & \multicolumn{2}{c}{Physics} & \multicolumn{4}{c}{Visual \& Semantics}
 & \multicolumn{2}{c}{Physics} & \multicolumn{4}{c}{Visual \& Semantics} \\
\cmidrule(lr){2-3} \cmidrule(lr){4-7} \cmidrule(lr){8-9} \cmidrule(lr){10-13} \cmidrule(lr){14-15} \cmidrule(lr){16-19} \cmidrule(lr){20-21} \cmidrule(lr){22-25}

 & OBR$\downarrow$ & CNR$\downarrow$ & Ra.$\uparrow$ & Spa.$\uparrow$ & Ac.$\uparrow$ & \textbf{Avg.}$\uparrow$
 & OBR$\downarrow$ & CNR$\downarrow$ & Ra.$\uparrow$ & Spa.$\uparrow$ & Ac.$\uparrow$ & \textbf{Avg.}$\uparrow$
 & OBR$\downarrow$ & CNR$\downarrow$ & Ra.$\uparrow$ & Spa.$\uparrow$ & Ac.$\uparrow$ & \textbf{Avg.}$\uparrow$
 & OBR$\downarrow$ & CNR$\downarrow$ & Ra.$\uparrow$ & Spa.$\uparrow$ & Ac.$\uparrow$ & \textbf{Avg.}$\uparrow$ \\
\midrule


DiffuScene~\cite{tang2024diffuscene} 
& 50.8 & \underline{30.3} & \underline{7.5} & \underline{7.3} & 8.5 & \underline{7.8}
& 62.8 & \underline{24.1} & \underline{7.4} & \underline{7.0} & \textbf{8.7} & \underline{7.7} 
& 27.8 & \underline{30.2} & \underline{7.3} & \underline{7.0} & 7.7 & \underline{7.3} 
& 47.1 & \underline{28.2} & \underline{7.4} & \underline{7.1} & 8.3 & \underline{7.6} \\

LayoutVLM~\cite{sun2025layoutvlm} 
& \underline{10.0} & 39.7 & 6.7 & 6.3 & \underline{8.7} & 7.2 
& \underline{8.1} & 63.1 & 7.2 & 6.6 & 8.3 & 7.4 
& \underline{6.5} & 31.0 & 6.5 & 6.4 & \textbf{8.7} & 7.2 
& \underline{8.2} & 44.6 & 6.8 & 6.4 & \underline{8.6} & 7.3 \\

\midrule

\textbf{Ours} 
& \textbf{1.1} & \textbf{1.8} & \textbf{8.8} & \textbf{8.1} & \textbf{8.9} & \textbf{8.6}
& \textbf{2.7} & \textbf{1.3} & \textbf{7.7} & \textbf{7.9} & \underline{8.5} & \textbf{8.0} 
& \textbf{1.1} & \textbf{1.0} & \textbf{7.9} & \textbf{7.8} & \textbf{8.7} & \textbf{8.1}
& \textbf{1.6} & \textbf{1.4} & \textbf{8.1} & \textbf{7.9} & \textbf{8.7} & \textbf{8.2} \\

\bottomrule
\end{tabular}%
}
\end{table*}

\begin{figure*}[!t]
  \centering
  \includegraphics[width=\textwidth]{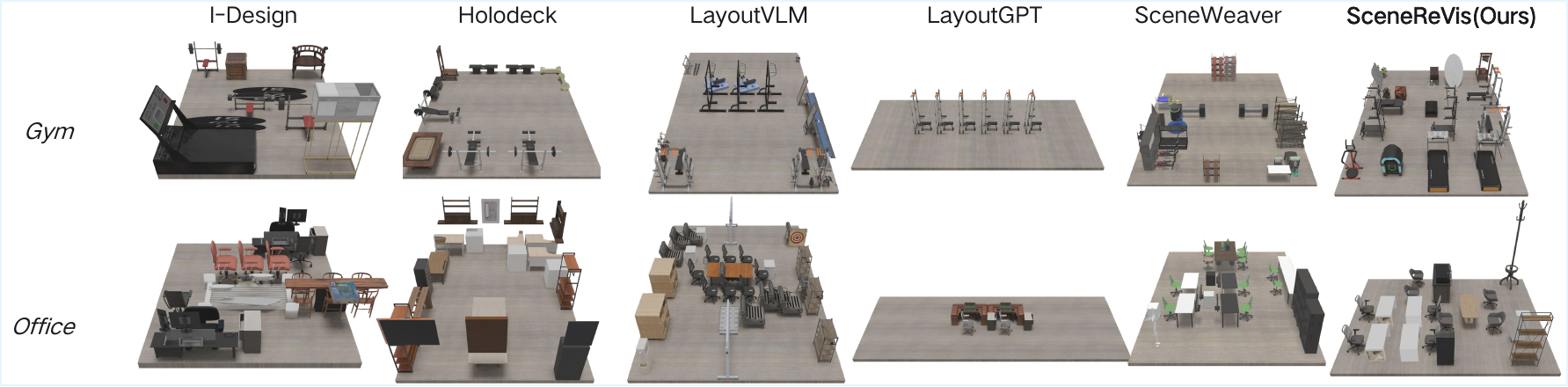}
  \caption{\textbf{Qualitative comparisons on additional OOD categories.}}
  \Description{Additional qualitative comparisons of generated scenes across methods.}
  \label{fig:more_com}
\end{figure*}

\begin{figure}[!t]
  \centering
  \includegraphics[width=0.95\columnwidth]{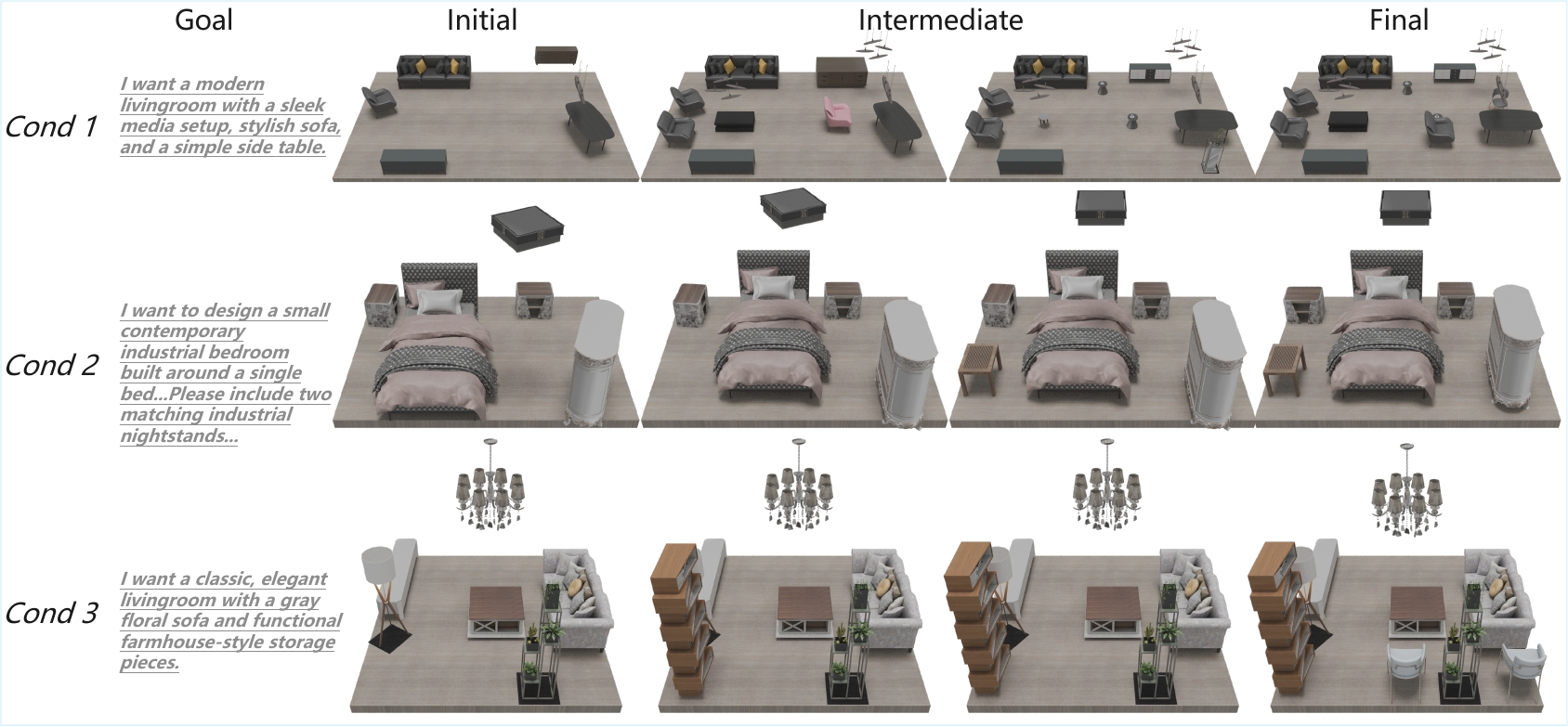}
  \caption{\textbf{Qualitative results for goal-oriented scene optimization.} We show trajectories under \textit{Chaotic \& Missing}, \textit{Chaotic Only}, and \textit{Missing Only} conditions.}
  \Description{Optimization trajectories for three goal-oriented scene optimization settings.}
  \label{fig:result_opt}
\end{figure}

\subsection{Performance on Standard Scene Generation}
Table~\ref{tab:comparison_bed_living} shows that SceneReVis achieves the highest visual and semantic score (Avg.=8.8) and the best CNR. Holodeck benefits from rule-based post-processing but can produce crowding, yielding lower accessibility than ours (8.7 vs. 8.9), while data-driven and pure-LLM baselines exhibit more physical violations. We attribute this balance to two-stage alignment for spatial diagnosis and planning, followed by physical optimization for fine-grained validity. Figure~\ref{fig:qualitative_standard} confirms that SceneReVis avoids severe collisions and unnatural placements while preserving semantic quality.

\subsection{Performance on Long-Tail and Additional OOD Scenarios}
Table~\ref{tab:comparison_dining_study_avg} evaluates underrepresented room categories. Following SceneWeaver~\cite{yangsceneweaver}, we exclude data-driven baselines~\cite{tang2024diffuscene,bucher2025respace} with incomplete category coverage. SceneReVis maintains strong physical and semantic quality, with the best CNR on Study and Avg.; SceneWeaver obtains lower OBR but weaker semantic quality, while LayoutGPT and LayoutVLM remain prone to collisions. This suggests that our policy transfers planning behavior rather than memorizing dominant categories. Figures~\ref{fig:qualitative_ood} and~\ref{fig:more_com} show similarly coherent layouts across long-tail and additional OOD categories. More results are provided in the supplementary material.

\begin{table}[t]
\centering
\caption{\textbf{Ablation study of key components.} VBL measures the volume of physical violations, and Phy. is the mean of OBR and CNR. Final Reward is the terminal reward, while Trajectory Reward is accumulated over multiple turns.}
\label{tab:ablation_component_wise}

\setlength{\tabcolsep}{1.8pt}

\resizebox{\columnwidth}{!}{%
\begin{tabular}{cccc | cc cccc}
\toprule

\multicolumn{4}{c|}{\multirow{2}{*}{\textbf{Components}}} & \multicolumn{6}{c}{\textbf{Metrics}} \\
\cmidrule(l){5-10}

\multicolumn{4}{c|}{} & \multicolumn{2}{c}{\textbf{Physics}} & \multicolumn{4}{c}{\textbf{Visual \& Semantics}} \\
\cmidrule(lr){1-4} \cmidrule(l){5-6} \cmidrule(l){7-10}

SFT & Final Reward & Trajectory Reward & Physical Opt. & VBL$\downarrow$ & Phy.$\downarrow$ & Ra.$\uparrow$ & Spa.$\uparrow$ & Ac.$\uparrow$ & \textbf{Avg.}$\uparrow$ \\

\midrule

\checkmark & - & - & -
& 544.5 & 30.4 & 8.7 & 7.1 & 7.8 & 7.9 \\

\checkmark & \checkmark & - & -
& 154.0 & 15.8 & 7.7 & 7.1 & 8.9 & 7.9 \\

\checkmark & \checkmark & \checkmark & -
& 210.9 & 21.9 & 9.1 & 7.7 & 8.7 & 8.5 \\

\midrule

\checkmark & \checkmark & \checkmark & \checkmark
& \textbf{25.6} & \textbf{3.3} & \textbf{9.4} & \textbf{8.0} & \textbf{8.9} & \textbf{8.8} \\

\bottomrule
\end{tabular}%
}
\end{table}

\subsection{Performance on Goal-oriented Scene Optimization} Table~\ref{tab:goal_oriented_all_in_one} shows that SceneReVis is robust across all three initial states, achieving near-perfect physical alignment, minimal boundary violations, and the best semantic quality. DiffuScene exhibits severe boundary violations, while LayoutVLM suffers collisions under clutter, highlighting how our diagnostic reasoning corrects specific scene defects.

\begin{table}[t]
\centering
\caption{\textbf{Human-ranking user study} (lower is better).}
\label{tab:user_study}
\setlength{\tabcolsep}{5pt}
\resizebox{\columnwidth}{!}{%
\begin{tabular}{l cc}
\toprule
\textbf{Method} & \textbf{Physical Plausibility}$\downarrow$ & \textbf{Visual \& Semantic Quality}$\downarrow$ \\
\midrule
I-Design~\cite{ccelen2024design}  & 4.63 $\pm$ 1.25 & 4.73 $\pm$ 1.35 \\
LayoutVLM~\cite{sun2025layoutvlm} & 4.30 $\pm$ 1.39 & 4.15 $\pm$ 1.51 \\
LayoutGPT~\cite{feng2023layoutgpt} & 3.97 $\pm$ 1.59 & 3.98 $\pm$ 1.54 \\
Holodeck~\cite{yang2024holodeck}  & 3.18 $\pm$ 1.71 & 3.21 $\pm$ 1.60 \\
SceneWeaver~\cite{yangsceneweaver} & \underline{2.71 $\pm$ 1.37} & \underline{2.92 $\pm$ 1.52} \\
\midrule
\textbf{SceneReVis (Ours)} & \textbf{2.20 $\pm$ 1.46} & \textbf{2.01 $\pm$ 1.15} \\
\bottomrule
\end{tabular}%
}
\end{table}

\subsection{Ablation and User Studies}
Table~\ref{tab:ablation_component_wise} shows that SFT alone leaves substantial physical violations. The Final Reward improves geometric validity, while the dense Trajectory Reward guides intermediate decisions and improves semantic quality. Physical optimization then removes residual conflicts overlooked by the VLM, yielding the strongest full-model results. In a study with 23 Master's/Ph.D. students and faculty members experienced in AI and computer graphics, SceneReVis achieves the best mean rank for both physical plausibility and visual-semantic quality (Table~\ref{tab:user_study}), confirming the quantitative improvements.
\section{Conclusion}
In this paper, we introduced \textbf{SceneReVis}, a self-reflective, vision-grounded framework that redefines 3D synthesis as a multi-turn reasoning process. Empowered by our \textbf{SceneChain-12k} dataset and a two-stage Agentic RL training strategy, the agent leverages iterative visual diagnosis to resolve complex geometric constraints. Extensive evaluations demonstrate state-of-the-art quality on mainstream categories and strong generalization on in-dataset underrepresented categories, while exploratory qualitative results suggest promising extension to additional OOD room types.

\begin{acks}
This work was completed during internships at Microsoft Research Asia (MSRA).

This work was supported by the Key Technology Research and Development Program of Shanghai (25JC3202303).
\end{acks}

\bibliographystyle{ACM-Reference-Format}
\balance
\bibliography{main}

\end{document}